\pdfoutput=1

\documentclass[11pt]{article}

\usepackage{EMNLP2022}

\usepackage{times}
\usepackage{latexsym}
\usepackage{multirow}
\usepackage{comment}
\usepackage{multicol}
\usepackage{multirow}
\usepackage{booktabs}
\usepackage{xspace}
\usepackage{cleveref}
\usepackage{tabularx}
\usepackage[linewidth=1pt]{mdframed}

\usepackage[T1]{fontenc}

\usepackage[utf8]{inputenc}

\usepackage{microtype}

\usepackage{inconsolata}

\newcommand{\reminder}[1]{[#1]}
\newcommand{\corpus}{\textsc{CoQASum}}
\newcommand{\task}{CQA summarization}
\newcommand{\model}{DedupLED}

\definecolor{c1}{HTML}{4e79a7}%
\definecolor{c2}{HTML}{f28e2b}%
\definecolor{c3}{HTML}{009E73}%
\definecolor{c4}{HTML}{56B4E9}%
\definecolor{c5}{HTML}{CC79A7}%
\definecolor{c6}{HTML}{E69F00}%
\definecolor{c7}{HTML}{844E4D}%
\definecolor{c8}{HTML}{2D512A}%

\newcommand{\hll}[1]{{\color{blue} #1}}
\newcommand{\tingyao}[1]{\reminder{{\bf\small\color{purple} (Ting-Yao)~#1}}}
\newcommand{\xiaolan}[1]{\reminder{{\small\color{orange} (Xiaolan)~#1}}}

\newcommand{\todo}[1]{\reminder{{\bf\small\color{red} (TODO)~#1}}}

\usepackage{tikz}
 
\title{Summarizing Community-based Question-Answer Pairs}

\author{Ting-Yao Hsu\thanks{~~Work done while at Megagon Labs.}\\
Pennsylvania State University\\
  \texttt{txh357@psu.edu} \\\And
Yoshi Suhara$^*$ \\
Grammarly\\
\texttt{yoshi.suhara@grammarly.com}\\\And
Xiaolan Wang$^*$ \\
Meta AI\\
\texttt{xiaolan@meta.com}
}

\begin{document}
\maketitle
\begin{abstract}

Community-based Question Answering (CQA), which allows users to acquire their desired information, has increasingly become an essential component of online services in various domains such as E-commerce, travel, and dining. However, an overwhelming number of CQA pairs makes it difficult for users without particular intent to find useful information spread over CQA pairs. To help users quickly digest the key information, we propose the novel \task\ task that aims to create a concise summary from CQA pairs. To this end, we first design a multi-stage data annotation process and create a benchmark dataset, \corpus, based on the Amazon QA corpus. We then compare a collection of extractive and abstractive summarization methods and establish a strong baseline approach DedupLED for the \task\ task. Our experiment further confirms two key challenges, sentence-type transfer and deduplication removal, towards the \task\ task. Our data and code are publicly available.\footnote{\url{https://github.com/megagonlabs/qa-summarization}}



\end{abstract}

\section{Introduction\label{sec:introduction}}
\begin{figure}[t]
\includegraphics[width=0.49\textwidth]{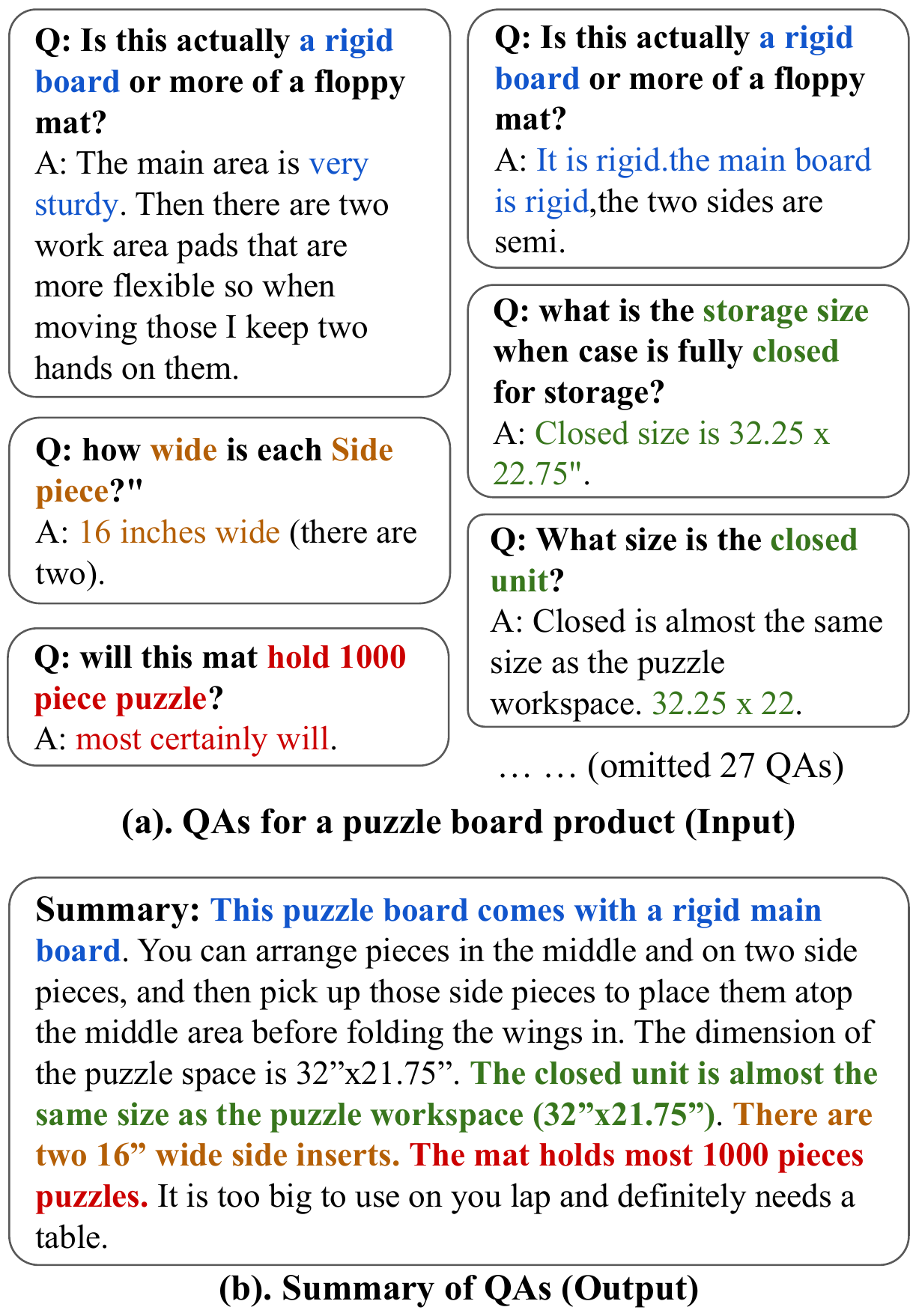}
    \caption{Example of the \task\ task. The input contains a collection of QA pairs. Duplicated information can be found in a single QA pair or across multiple QA pairs. The output is a concise and coherent summary written in declarative sentences.  }
    \label{fig:amazon_qa}
\label{fig:example}
\end{figure}

Community-based Question Answering (CQA) enables users to post their questions and obtain answers from other users. With the increase in online services, CQA has become essential for various purposes, including online shopping, hotel/restaurant booking, and job searching. Many online platforms implement CQA features to help users acquire additional information about entities (e.g., products, hotels, restaurants, and companies) of their interests.
CQA complements customer reviews---another type of user-generated content, which provide additional information about the entity but mostly focusing on user experiences and their opinions. 

While CQA greatly benefits users in decision-making, digesting information from original question and answer pairs (QA pairs\footnote{In this paper, we use QA pairs to refer to question-answer pairs in CQA.}) also has become increasingly harder. 
Due to the community-based nature, CQA tends to have a large number of heavily repetitive QA pairs, 
which make it difficult for users, especially those who do not have specific intent (i.e., questions), to find and digest key information. 

Existing summarization efforts for CQA~\cite{liu2008understanding,Deng:2020:AAAI,AnswerSumm} primarily focus on summarizing answers for a given question, which still assumes that the user has a certain intent.
We believe that information spread over QA pairs can be summarized into a more concise text, which helps any users grasp the key points of discussions about a target entity.
Therefore, we take a step beyond the scope of answer summarization and propose a novel task of \task, which aims to summarize a collection of QA pairs about a single entity into a concise summary in declarative sentences (shown in Figure~\ref{fig:amazon_qa}).

%

The \task{} task has the following two unique challenges.
%
First, \task{} needs to solve sentence-type transfer 
as questions in interrogative sentences have to be converted into declarative sentences to make a concise summary. This challenge is not trivial as existing summarization tasks assume that input and output are both written in declarative sentences. 
%
Second, CQA contains duplicated questions and answers. That is, different users can post similar questions. A question can have multiple answers, many of which contain duplicate information. 
Also, unlike question-answering forums (e.g., Quora), CQA in online services is less incentivized to remove such redundancy. Slightly different questions/answers can provide detailed and useful information not mentioned in other questions/answers. Having more similar answers supports the information is more reliable.
Those properties make existing summarization solutions unsuitable for \task. 

To enable further study of the \task\ task, we create a corpus \corpus{} by collecting reference summaries on QA pairs from the Amazon QA dataset \cite{wan2016modeling, mcauley2016addressing}.
Reference summary annotation is challenging for \task, as a single entity (i.e., a product for the dataset) can have so many questions and answers that the annotator cannot write a summary directly from them. Furthermore, the sentence-type difference (i.e., interrogative vs. declarative) obstructs summary writing. 
To make this annotation feasible, we designed a multi-stage annotation framework. Sampled seed QA pairs are given to the annotator to convert into declarative sentences, which are then rewritten into gold-standard summaries by expert writers. At the last step, we collected semantically similar QA pairs to to make the annotated corpus more realistic.

We conduct a comprehensive experiment that compares a collection of extractive and abstractive summarization solutions and establish a strong baseline approach, DedupLED, for the \task\ task. Specifically, \model{} fine-tunes the entire LED model for summary generation while additional classifier attached to the encoder is optimized to extract representative QA pairs. Leveraging the strengths of both abstractive and extractive summarization objectives, as well as the pre-trained language model checkpoints, DedupLED significantly outperforms the other alternative methods. 
Our experiment also confirms that \model{} is suitable for \task, as the model implements the functionality for both (1) sentence-type transfer and (2) duplicate removal.

Our contributions of the paper are as follows:
\begin{itemize}
    \setlength{\parskip}{0cm}
    \setlength{\itemsep}{0cm}
    \item We propose the novel task of \task, which takes QA pairs about a single entity as input and make a summary written in declarative sentences (Section~\ref{sec:problem}). 
    \item We designed a multi-stage annotation framework and collected reference summaries to build the first benchmark corpus for \task. The corpus is based on the Amazon QA corpus~\cite{wan2016modeling} and consists of reference summaries for $1,440$ entities with $39,485$ QA pairs from 17
    product categories. (Section~\ref{sec:corpus}). 
    \item We conduct comprehensive experiments on a collection of extractative and abstractive summarization methods and develop a strong baseline \model, which implements key characteristics of sentence-type transfer and duplication removal functions. 
    (Section~\ref{sec:models} and Section~\ref{sec:evaluation}). 
\end{itemize}

\section{Problem definition\label{sec:problem}}
Let $D$ denote a dataset of questions and answers on individual entities $\{e_{1}, e_{2},...,e_{\left|D \right|}\}$ (e.g., products or hotels).
For every entity $e$, we define a set of question-answer pairs $QA_e=\{(q_i, a_i)\}^{|QA_e|}_{i=1}$, where the question $q_i$ and the answer $a_i$ are sequences of tokens $q_i = (w_1,...,w_n)$ and $a_i = (a_1, ..., a_m)$\footnote{Note that one question can have multiple answers, but we use this ``flat'' notation for simplicity. Thus, there can exist $q_i = q_j$ for some $i \ne j$. 
}.
Given a set of QA pairs for an entity $e$, the \task{} task is 
to generate a natural language summary $S_{e}$ from $QA_e$.

\begin{figure*}[th]
    \centering
    \includegraphics[width=0.92\textwidth]{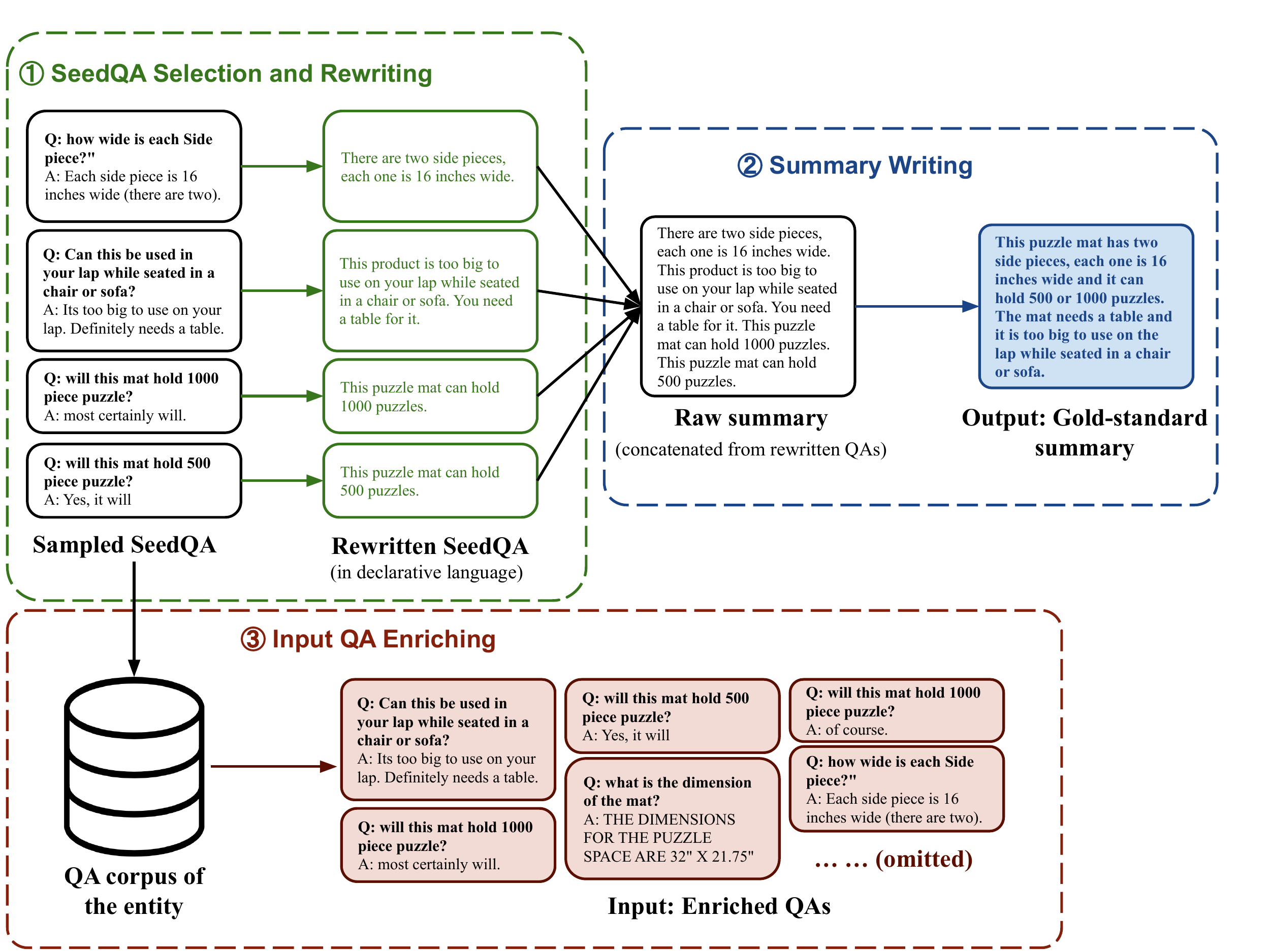}
    \caption{Overview of our multi-stage annotation framework for \task. 
    }
    \label{fig:annotation}
\end{figure*}

\section{The \corpus{} Corpus\label{sec:corpus}}

\begin{table*}[h]
\small
\centering
\begin{tabular}{l|r|rrr|rrrr}
\toprule
\multirow{2}{*}{\textbf{Category}} & \multicolumn{1}{c|}{\multirow{2}{*}{\textbf{|Entity|}}} & 
\multicolumn{3}{c|}{\textbf{Avg. word count}} &
\multicolumn{4}{c}{\textbf{\% of novel n-grams in gold summary}}
\\ 
     &  & 
      \multicolumn{1}{c}{\textbf{Input}} & 
      \multicolumn{1}{c}{\textbf{Raw sum.}} &
      \multicolumn{1}{c|}{\textbf{Ref sum.}} &
      \multicolumn{1}{c}{\textbf{unigram}} &
      \multicolumn{1}{c}{\textbf{bigram}} &
      \multicolumn{1}{c}{\textbf{trigram}} & 
      \multicolumn{1}{c}{\textbf{4-gram}}
     \\ \midrule
Automotive & 95 & 1044.8 & 167.1 & 117.3 & 21.16&57.80&75.43&83.37 \\
Baby       & 17 & 1162.7 & 156.4 & 113.7 & 28.56&69.56&86.09&92.69 \\
Beauty     & 20 & 1038.5 & 161.2 & 127.2 &23.96&62.84&81.82&89.15 \\
Cell Phones and Accessories & 94 & 931.5 & 135.8 & 96.5 & 16.24&50.56&69.07&78.18 \\
Clothing Shoes and Jewelry & 10 & 1134.8 & 159.3 &130.2 & 16.67&48.17&64.81&72.92\\
Electronics & 304 & 1000.6 & 167.3 & 132.1 & 22.06&55.11&70.73&77.77 \\
Grocery and Gourmet Food & 22 & 908.4 & 123.4 & 96.7 &16.17&54.91&74.52&82.81 \\
Health and Personal Care & 80 & 1125.7 & 139.6 & 103.6 &15.42&52.75&71.10&79.74 \\
Home and Kitchen & 262 & 1093.6 & 153.2 & 113.5 & 23.00&62.04&79.08&86.21 \\
Musical Instruments & 22 & 900.0 & 197.0 & 142.1 & 35.06&66.65&78.40&82.85 \\
Office Products & 66 & 994.3 & 141.4 & 103.7 & 12.69&45.20&63.21&72.94 \\
Patio Lawn and Garden & 70 & 1177.2 & 142.1 &107.5 & 12.92&47.08&65.16&74.80 \\
Pet Supplies & 11 & 1154.9 & 124.8 & 106.5 & 16.51&53.42&73.91&83.84\\
Sports and Outdoors &163 & 1120.0 & 143.7 &106.4 &13.85&47.57&66.26&75.65 \\
Tools and Home Improvement & 138&1096.0 & 167.6 & 110.0 &18.68&45.58&58.63&65.07\\
Toys and Games & 54 & 984.7 & 150.2 & 112.3 &21.41&60.85&79.28&86.94 \\
Video Games & 12 & 1087.6 & 186.4 & 128.7 & 28.59&61.22&76.87&83.70 \\
\midrule
\textbf{All} & 1,440 & 1055.6 & 154.8 & 114.8& 19.46&54.17&71.06&78.88
\\ \bottomrule
\end{tabular}
 \caption{\corpus{} 
 dataset statistics. 
 }
\label{tab:qa_dataset-stats}
\end{table*}

We first describe the multi-stage annotation framework to collect gold-standard reference summaries from input QA pairs and then describe our benchmark dataset \corpus.

\subsection{A Multi-stage Annotation Framework}\label{sec:annotation}
Reading and summarizing a set of QA pairs is challenging and error-prone for three reasons: (1) a large number of QA pairs, (2) the heavy repetition and noise in both questions answers, and (3) the difficulty of converting questions and answers into declarative summaries.
Thus, it is infeasible to collect high-quality reference summaries by simply showing a set of QA pairs and asking annotators to write a summary.
In this paper, we design a multi-stage annotation framework that first simplifies this complex annotation task into more straightforward annotation tasks and then enriches the collected annotations.

Figure~\ref{fig:annotation} depicts the schematic procedure of the multi-stage annotation framework.
For each entity and its corresponding QA pairs in the original corpus, we first select representative seed QA pairs and ask annotators to rewrite them into declarative sentences, which are then concatenated into a raw summary. 
Next, we ask highly-skilled annotators to polish the raw summary into a more fluent summary. 
In the last step, we enrich the seed QA pairs by selecting semantically similar QA pairs from the original corpus.


\subsubsection*{Step 1: QA Pair Selection and Rewriting}
In this step, we use a drastic strategy to remove duplicate QA pairs and simplify the annotation task for human annotators. 
A natural way to deduplicate QA pairs is by manually comparing existing QA pairs' semantics and only keeping the unique ones. However, we found this approach less practical because asking human annotators to perform the comparison is extremely expensive. It is also hard to validate the quality because selecting a representative QA from a set of semantically similar ones is a subjective process.

Thus, we use a heuristic-based strategy to select representative QA pairs from the original corpus. 
Specifically, we use the following two rules to filter out QA pairs that are not suitable for creating reference summaries: (1) \textit{length rule}: QA pairs with less than $5$ or more than $150$ tokens;
(2) \textit{pronoun rule}: QA pairs that include first-person pronouns. We found that long questions/answers tend to contain their background information (e.g., personal stories), which is irrelevant to the entity. First-person pronouns are also a strong indicator for such questions/answers.
After the filtering, we randomly sample $k$ seed QA pairs from the remaining ones. In addition, to avoid redundancy, we only sample seed QA pairs of different questions.\footnote{Note that there is a chance that selected QA pairs contain duplicate information. We make sure to exclude such duplicate information in Step 2.}

For each of the $k$ seed QA pairs, we ask human annotators to rewrite them into declarative sentences. We recruited three crowd workers from Amazon Mechanical Turk\footnote{\url{https://www.mturk.com/}} to annotate every QA pair and chose the highest-quality annotation based on 6 criteria: (1) length of LCS against the original QA pair, (2) use of yes/no, (3) use of interrogative determiner (e.g., What), (4) use of first-person pronouns, (5) use of the item name at the beginning, (6) the ignorance of the question information. We also blocked MTurk workers with consistently low-quality annotations to ensure the quality of annotations.
%


\subsubsection*{Step 2: Summary Writing}
We form a raw summary by concatenating annotations (i.e., declarative sentences rewritten from QA pairs) obtained in the first step for each entity.
The raw summaries are not necessarily fluent and coherent as different pieces are annotated independently. They may also contain redundant information. 
To address these issues, we use another annotation task to polish and write a summary from the raw summary. To ensure the quality, we hired highly-skilled writers from  Upwork\footnote{\url{https://www.upwork.com/}} by conducting screening interviews for this annotation task.
For each entity, we show annotators the raw summary and ask them to write a fluent and concise summary. 

\subsubsection*{Step 3: Enriching Input QA Pairs}
Recall that in Step 1, we select $k$ seed QA pairs for each entity. The seed QA pairs are less redundant because of the filtering and sampling strategy. This does not reflect the real-world scenario, where similar questions are asked multiple times, and each question often contains several answers.

To align the benchmark with more realistic settings, we enrich the input QA pairs in Step 3. 
In particular, we add all answers to every question in the seed QA pairs. Besides, we retrieve questions that are semantically similar to seed questions and add all the answers to the input QA pairs. For semantic similarity calculation, we use BERT embeddings and word overlap to find the candidates, followed by an additional crowd-sourcing task using Appen\footnote{\url{https://appen.com/}} for manual validation.
The validation step ensures that our reference summaries can be created from the enriched input QA pairs.

\subsection{Dataset Statistics}
Using the multi-stage annotation framework, we created the \corpus\ benchmark based on the Amazon QA dataset~\cite{wan2016modeling,mcauley2016addressing}.
We selected 1,440 entities from 17 product categories with 39,485 input QA pairs and 1,440 reference summaries. Besides, \corpus{} also contains rewritten QA pairs in declarative sentences for the QA pair rewriting task in Step 1, which consist of 3 annotations for each of the 11,520 seed QA pairs ($k=8$ seed QA pairs for each entity). 






Table~\ref{tab:qa_dataset-stats} shows the statistics of \corpus. 
We confirm that the average word count of input QA pairs/raw summaries/reference summaries is consistent for different categories.
%
The novel n-gram distributions also confirm that \corpus{} offers a fairly abstractive summarization task. Some product categories such as ``Office Products'' and ``Patio Lawn and Garden'' have lower novel n-gram ratios, indicating that the task becomes relatively extractive. 
%
The word count difference between the raw summary and the reference summary supports the value and quality of the summary writing task in Step 2, indicating that the raw summary still contains some redundant information.

\section{Models\label{sec:models}}
To examine the feasibility and explore the challenges of \task, we tested several summarization solutions on \corpus. The models are grouped into {\em Extractive}, {\em Extractive-Abstractive} and {\em Abstractive} methods. 



\subsection{Extractive}
%
Extractive methods extract salient sentences from input QA pairs as the output summary. We consider unsupervised (LexRank) and supervised (BertSumExt) models in addition to a simple rule-based baseline that filters the original seed input QA. 
We evaluate those methods to understand how well selecting sentences without sentence-type transfer performs on the task.

\paragraph{SeedQAs:} This method concatenates the seed QA pairs used in the first annotation task of the multi-stage annotation framework. This is an oracle method as we cannot tell which QA pairs were used as seed QA pairs for annotation. We use this method to verify the performance of simply extracting QA pairs. 


\paragraph{LexRank~\cite{erkan2004lexrank}:} This is an unsupervised extractive method, which uses the similarity between words to build a sentence graph and compute the centrality of sentences for selecting top-ranked sentences as summary. 



\paragraph{BertSumExt~\cite{liu2019text}:} BertSumExt is a supervised model, which fine-tunes BERT~\cite{devlin2019bert} to extract sentences by solving multiple sentence-level classifications. 
In our experiment, we use BertSumExt to extract salient QA pairs from the input, where the gold-standard labels are acquired by greedily select QA pairs that maximize the ROUGE scores to the gold-standard summary\footnote{\url{https://github.com/nlpyang/BertSum}}.

\subsection{Extractive-Abstractive}
While extractive methods can remove duplication from the input, they cannot transfer interrogative sentences (i.e., questions) into declarative sentences.
To handle this better, we combine extractive and abstractive models to implement two-stage solutions. We also test an existing two-stage algorithm in addition to another summarization model that learns to extract and rewrite in an end-to-end manner. 


\paragraph{LexRank+LED} 
This method is a {\em select-then-rewrite} hybrid model. Using a sentence-type transfer model, the model rewrites each of the QA pairs extracted by LexRank into declarative sentences, which are then concatenated as an output summary. 
For the sentence-type transfer model, we fine-tune the LED model~\cite{beltagy2020longformer} on the seed QA pairs and their rewritten texts collected in Step 1 of the multi-stage annotation pipeline (Section~\ref{sec:annotation}). 

\paragraph{LED+LexRank} 
This method is a {\em rewrite-then-select} hybrid model that swaps the steps of LexRank+LED. It uses LexRank to extract salient sentences from input QA pairs rewritten by the same sentence-type transfer model.

\paragraph{Bert-SingPairMix~\cite{lebanoff2019scoring}} Bert-SingPairMix is a select-then-rewrite-style model that first selects salient sentences from the input and then summarizes the selected sentences into the summary. In our experiment, we use our gold-standard summaries to train both the content selection model and the abstractive summarizer. 


\paragraph{FastAbstractiveSum~\cite{chen2018fast}}
FastAbstractiveSum also implements select-then-rewrite summarization via reinforcement learning. The model learns to select representative sentences with the extractor and rewrite the selected sentences with the abstractor. We train a FastAbstractiveSum model on the gold-standard summaries.


\subsection{Abstractive}
As the final group of models, we explore abstractive models that directly summarize input QA pairs. Specifically, we use LED and its variants, which can take long-document as input. Our \model{} is a variant of LED and falls into this group.




\paragraph{LED~\cite{beltagy2020longformer}} 
This model fine-tunes Longformer Encoder-Decoder (LED)~\cite{beltagy2020longformer} on input QA pairs and the gold-standard summaries in the training set. 


\paragraph{HierLED~\cite{zhu2020hierarchical,zhang2021emailsum}}
Hierarchical LED (HierLED) is a variant of LED, which has two encoders for token-level and QA-level inputs to handle the structure of QA pairs better. 
We use the same architecture as Hierarchical T5~\cite{zhang2021emailsum}, replacing T5 with LED. We fine-tune the model in the same manner as LED.


\begin{figure}[t]
    \centering
    \includegraphics[width=0.45\textwidth]{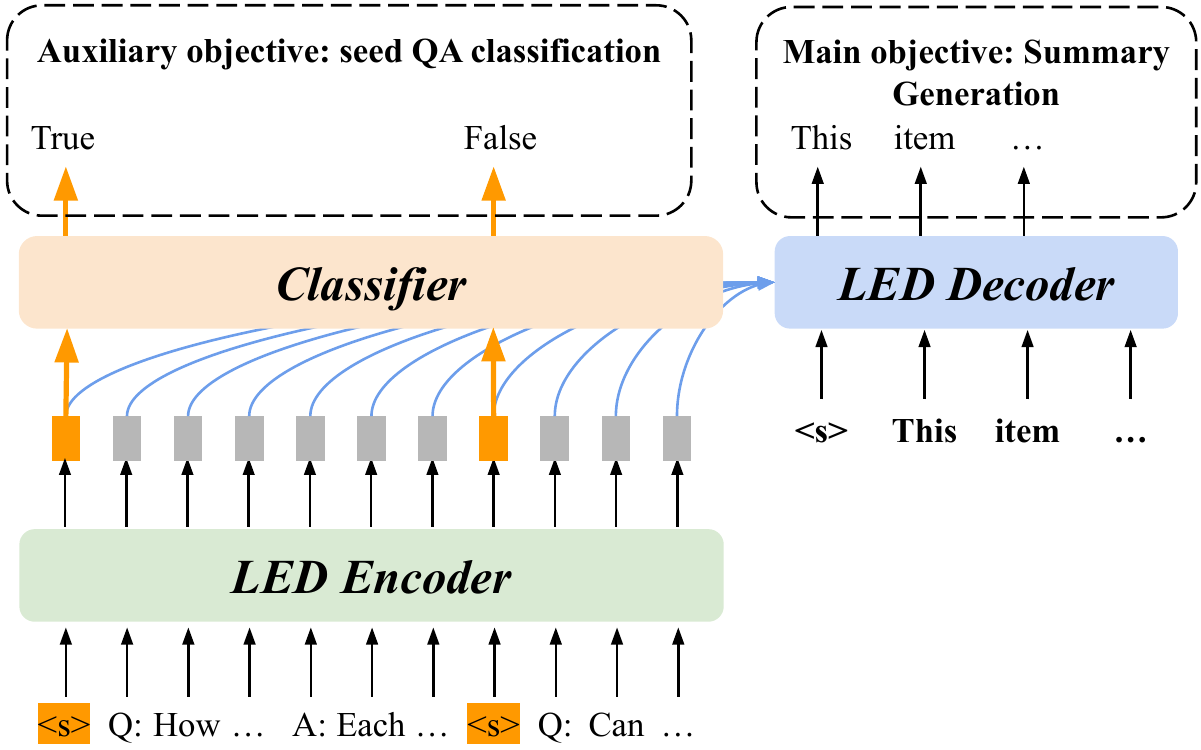}
    \caption{Architecture of \model.}
    \label{fig:dedupled}
    \vspace{-0.5cm}
\end{figure}


\paragraph{DedupLED} While pre-trained encoder-decoder models, including LED, are known to be powerful summarization solutions, they do not explicitly implement deduplication functionality. Inspired by BertSumExt, we consider incorporating a classifier layer optimized to extract the original seed QA pairs into an LED model and fine-tuning the LED model via multi-task learning, which we refer to as \model.
Figure~\ref{fig:dedupled} depicts the model architecture. The classifier layer is trained to select the original seed QA pairs, so the shared encoder learns to detect duplicate information while the decoder is optimized to generate a summary. 
In the training time, \model{} uses the original seed QA pair information in addition to the gold-standard summaries in the training data. We would like to note that \model{} does not require any additional information other than input QA pairs in the summary generation phase. 

\section{Evaluation\label{sec:evaluation}}
\begin{table*}[t]
    \centering
    \begin{tabular}{l|rrrr|ccc}
    \toprule  
    & \multicolumn{4}{c|}{\textbf{Performance}} & \multicolumn{3}{c}{\textbf{Supervision}} \\
     &\multicolumn{1}{c}{R1} &  \multicolumn{1}{c}{R2} &  \multicolumn{1}{c}{RL} &  \multicolumn{1}{c|}{BertS} & SeedQA & Rewr. QA & Gold Sum. \\ \midrule
    \textit{Extractive:} & &&&&\\
     \quad SeedQAs & 18.96 & 10.22 & 12.57 & 83.26 & - & - & -\\
     \quad LexRank & 33.17 & 9.30 & 19.26 & 83.76 & - & - & - \\
     \quad BertSumExt & 31.81 & 11.10 & 19.38 & 84.57 & No & No & Yes \\ \midrule
     \textit{Extractive-Abstractive:}& &&&&\\
    \quad LexRank+LED & 35.92 & 8.97 & 18.37 & 84.37 & No & Yes & No\\
    \quad LED+LexRank & 38.01 & 10.71 & 19.98 & 84.01 & No & Yes & No\\
    \quad BERT-SingPairMix & 40.82 & 12.73 & 21.28 & 85.17 & No & No & Yes \\
    \quad FastAbstractiveSum & 42.51 & 15.21 & 22.53 & 84.47 & No & No & Yes\\ \midrule

    
    
    \textit{Abstractive:} & &&&&\\
    \quad LED & 45.82 & 19.34 & 26.01 & 87.55  & No & No & Yes \\
    \quad  HierLED
    & 48.30 & 23.29 & 29.84 & 88.55 & No & No & Yes\\ 
    \quad \textbf{DedupLED} & \textbf{52.73} & \textbf{27.24} & \textbf{31.68} & \textbf{88.96} & Yes & No & Yes\\
     \bottomrule
    \end{tabular}
    \caption{
    Performance of the models on \corpus{} and the type of supervision that each method used. R1/R2/RL/BS denotes ROUGE-1/2/L F1 and BERTScore F1, respectively. With the auxiliary objective, \model{} outperforms all the other alternative models.}
    \label{tab:result-summarization}
\end{table*}

We conduct comparative experiments to evaluate those models for the \task{} task on the \corpus{} dataset.
We randomly split the data into train/validation/test sets, which consist of 1152/144/144 entities, respectively. 
For LexRank, we limit the output length based on the average reference summary length in the training set. For LED and its variations, we fine-tuned the \texttt{allenai/led-base-16384} checkpoint using the Hugging Face Transformers library.~\footnote{\url{https://github.com/huggingface/transformers}} We report the performance of the best epoch (based on ROUGE-1 F1) chosen on the validation set for all the supervised models.






\subsection{Automatic Evaluation}
For automatic evaluation, we use ROUGE~\cite{lin2004rouge} F1\footnote{\url{https://pypi.org/project/py-rouge/}} and BERTScore~\cite{zhang2019bertscore} F1\footnote{\url{https://github.com/Tiiiger/bert_score}} with the default configuration. 
The performance and required supervision of all models described in Section~\ref{sec:models} are shown in Table~\ref{tab:result-summarization}.

\paragraph{Extractive:}
SeedQAs, which simply selects the original QA pairs, performs badly. This is expected because while with high recall ($88.45$ R1-recall), the Oracle method suffers badly from low precision, largely due to the sentence-type inconsistency (i.e., interrogative vs. declarative) and duplication in input QA pairs. 
LexRank, the unsupervised summarization baseline, performs slightly better than SeedQAs thanks to its ability to select more concise QAs for the output summary. BertSumExt, while leveraging gold-standard summaries, achieves similar performance with LexRank. We believe the discrepancy between interrogative and declarative sentences in input QA pairs and gold-standard summaries is the primary cause of the performance. 



\paragraph{Extractive-Abstractive:}
Extractive-abstractive models achieve better performance than extractive models. The sentence-type transfer helps LexRank+LED/LED+LexRank achieve a much higher R1 score while comparative R2/RL/BS scores against the original LexRank. This implies the limitation of sentence selection before/after sentence-type transfer. Also, the sentence-type transfer model was trained on seed QA pairs and their corresponding declarative sentences, not the gold-standard summaries. Thus, another factor may be the difference between the rewritten QA pairs and the gold-standard summaries. 

Both FastAbstractiveSum and BERT-SingPairMix, which are directly supervised by the gold-standard summaries, show significantly better performance than the extractive models.
The results confirm that those models can learn to perform both sentence-style transfer and duplication removal directly from gold-standard summaries.


\paragraph{Abstractive:} All three models achieve strong performance on the \task\ task. The vanilla LED outperforms extractive/extractive-abstractive models. 
By incorporating the hierarchical structure into the model, HierLED improves the performance against the vanilla LED.
Furthermore, \model{} achieves the best performance for all the evaluation metrics. This confirms that by adding an auxiliary objective and using another supervision (i.e., seed QA pair selection), \model{} appropriately learns to deduplicate while learning to summarize input QA pairs. 


\noindent {\bf Takeaway:}
\textit{From the results, we confirm that both sentence-style transfer and duplication removal are crucial for the \task\ task. In addition, fine-tuning pre-trained language models using the gold-standard summaries offers strong performance, better than manually-crafted two-stage summarization models. Finally, by incorporating the duplication removal functionality into the model via multi-task learning, we show that \model{} establishes a strong baseline for the \task\ task.}



\subsection{Human evaluation}
We further conducted human evaluation to judge the quality of generated summaries by different models.
For every entity in the test set, we showed summaries generated by four models (LexRank, FastAbstractiveSum, BERT-SinglePairMix, and DedupLED) to three human judges\footnote{\url{https://appen.com/}} to choose the \textit{best} and \textit{worst} summaries for three criteria: informativeness (Inf.), coherence (Coh.), and conciseness (Con.). Then, we computed the performance of the models using the Best-Worst Scaling~\cite{louviere2015best}.
Table~\ref{tab:human_eval} shows that \model{} consistently achieves the best performance in all three criteria. On the other hand, LexRank, as expected, performs the worst among all the methods we tested.
The human evaluation performance trend aligns with the automatic evaluation performance, validating the quality of \corpus\ as a benchmark for the \task\ task.


\begin{table}[]
\small
    \centering
    \begin{tabular}{l|c|c|c}
    \toprule
    & Inf. & Coh. & Con. \\ \midrule
    LexRank &-8.41&-5.72&-8.42 \\
    FastAbstractiveSum &-4.71&+1.01&+3.37 \\
    BERT-SingPairMix &-2.02&-6.07&-1.35 \\
    DedupLED &+13.13&+5.72&+4.38\\\bottomrule
    \end{tabular}
    \caption{Best-Worst Scaling on human evaluation.}
    \label{tab:human_eval}
\end{table}

\section{Analysis\label{sec:analysis}}

\subsection{Choice of Pre-trained Language Models}
To justify our observation that pre-trained language models have strong abilities 
we test and compare three additional pre-trained language models on \corpus: PEGASUS~\cite{zhang2020pegasus}, T5~\cite{raffel2020exploring}, and BART~\cite{lewis2019bart}.
We confirm that all models perform better than the extractive and extractive-abstractive models. While PEGASUS and T5 show similar ($24.81$ and $24.61$ RL, respectively), they are less effective than BART and LED ($26.89$ and $26.01$ RL, respectively).
\subsection{Learning Curve Analysis}
\begin{figure}[t]
    \centering
    \includegraphics[width=0.45\textwidth]{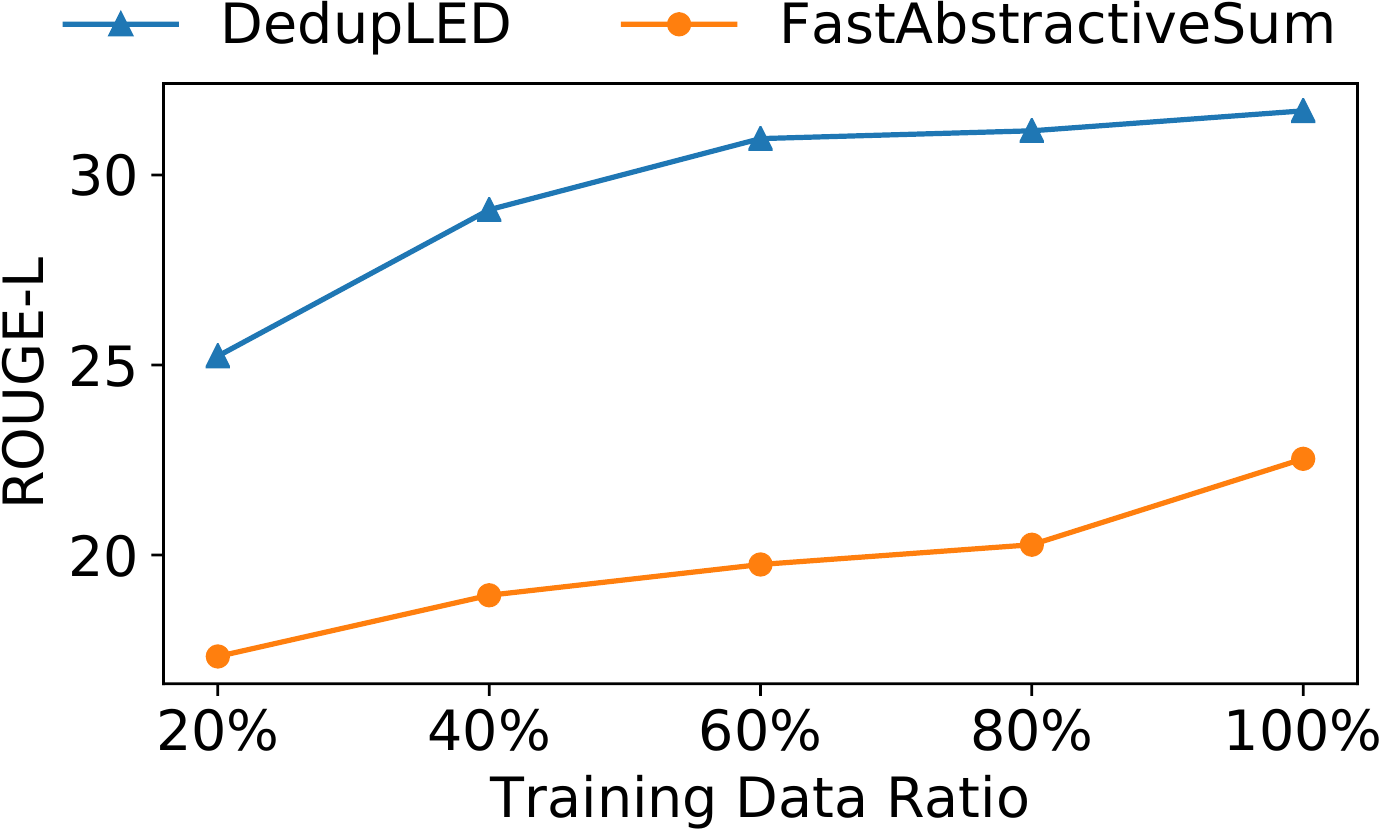}
    \caption{Learning curve analysis. $x$-axis is the training data ratio, and $y$-axis is the performance on the test set.}
    \label{fig:lowratio}
\end{figure}

Since collecting reference summaries is costly and time-consuming, we investigate the models' performance with limited training data. We tested the models' performance when trained with 20\%, 40\%, \dots, 100\% of the training data.
Figure~\ref{fig:lowratio} shows the ROUGE-L F1 scores of \model{} and FastAbstractiveSum when trained on different size of training data. 
By leveraging a pre-trained checkpoint, \model{} performs consistently and substantially better than FastAbstractiveSum, which is trained from scratch. 
\model{} also shows a faster learning curve and reaches the plateau in performance when trained with 60\% and more data. This supports that the annotation size of \corpus{} is sufficient for fine-tuning pre-trained language models, while it may be insufficient for non-pre-trained models.



\subsection{Cross-category Transfer Learning}
\corpus{} contains 17 different categories and varying amounts of entities within each category.
To investigate how different categories and numbers of training data affect the summarization performance, we experiment DedupLED on the top five categories in terms of entity count. We first fine-tuned DedupLED on each category and tested it on the five categories. For each category, we split entities into train/dev/test sets in 0.8/0.1/0.1 ratios.

Table~\ref{tab:cross-category2} shows ROUGE-1 F1 scores of the DedupLED models in a cross-category setting. 
We find that more training data generally helps improve the model quality even if not fine-tuned on training data in the same category. Electronics and Home \& Kitchen are the top two categories with the most training examples (243 and 209 entities, respectively), and achieved the best performance across all categories. 
%
From the results, we confirm that
summarization models based on pre-trained language models
have strong cross-category transfer ability in \task.

\begin{table*}[t]
    \centering
    \begin{tabular}{c|c|c|c|c|c|c|c|c}
    \toprule  
    &&\multicolumn{5}{c|}{{\bf Test category}} \\
    {\bf Training category} & {\bf |Train|}
    & \multicolumn{1}{c|}{ELEC}
    & \multicolumn{1}{c|}{H\&K}
    & \multicolumn{1}{c|}{S\&O}
    & \multicolumn{1}{c|}{T\&H}
    & \multicolumn{1}{c|}{AUTO}
    & \multicolumn{1}{c}{\bf Avg }\\\midrule
     \quad ELEC 
     & 243 
     & {\bf 46.75}
     & 41.74
     & {\bf 46.92}
     & 38.90
     & 40.52
     & \multicolumn{1}{c}{{\bf 42.97}} \\
     \quad H\&K 
     & 209 
     & 41.84
     & {\bf 41.97}
     & 44.46
     & 37.51
     & {\bf 42.53} 
     & \multicolumn{1}{c}{41.66}\\
     \quad S\&O
     & 130 
     & 38.26
     & 42.99 
     & 38.64
     & {\bf 39.17}
     & 36.18 
     & \multicolumn{1}{c}{39.05} \\
     \quad T\&H 
     & 110 
     & 42.86
     & 40.50 
     & 40.25
     & 38.40
     & 38.05 
     & \multicolumn{1}{c}{40.01} \\
     \quad AUTO 
     & 76 
     & 41.71 
     & 38.96
     & 40.34
     & 38.75
     & 37.53 
     & \multicolumn{1}{c}{39.46} \\
     \bottomrule
    \end{tabular}
    \caption{Cross-category performance (ROUGE-1 F1) of \model{} on top-5 product categories: Electronics (ELEC), Home and Kitchen (H\&K), Sports and Outdoors (S\&O), Tools and Home (T\&H), and Automatic (AUTO). Each model was trained on training data of each category and then tested on the five categories. The highest ROUGE-1 F1 scores for each category are bold-faced.}
    \label{tab:cross-category2}
\end{table*}

\section{Related Work\label{sec:related-work}}
Opinion summarization~\cite{OpinionSummarizationSIGIR2022Tutorial}
aims to create a summary from multiple customer reviews. While opinion summarization is relevant to \task{} as it summarizes consumer-generated text, customer reviews are significantly different from QA pairs in CQA as they are self-contained and tend to contain more subjective information. Recent opinion summarization models have adopted pre-trained LMs (LED) for summarizing multiple reviews~\cite{oved-levy-2021-pass,iso-etal-2022-comparative}. 

A line of work studies on summarizing answers in CQA, which can be categorized into extractive models~\cite{liu2008understanding,chan-etal-2012-community,Deng:2020:AAAI,Deng:2020:SIGIR} and abstractive models~\cite{AnswerSumm,chowdhury2021neural}.
Among them, \citet{CQASumm} created a benchmark by selecting the best answer as the reference summary, and \citet{AnswerSumm} has collected professionally written reference summaries for answer summarization.
Our \task{} differs from answer summarization as we consider multiple QAs as input, which offers unique challenges not in answer summarization.

Another line of work in dialog summarization has created new benchmarks for E-mail threads~\cite{zhang2021emailsum}, customer support conversations~\cite{feigenblat-etal-2021-tweetsumm-dialog}, conversations in multiple domains~\cite{fabbri-etal-2021-convosumm}, and forum discussions~\cite{khalman-etal-2021-forumsum-multi}.
\task{} is similar to those tasks in creating abstractive summaries from multiple turn-taking conversations between more than one user. Meanwhile, we also found that \task\ tends to contain more duplication in the input by nature as the compression ratio (i.e., input length/summary length) of \corpus\ is $10.88\%$, which is smaller than EmailSum ($29.38\%$) and ForumSum ($11.85\%$).
We also tested HierLED, a variant of the strongest baseline for E-mail thread summarization, and confirmed that \model{} performs better than HierLED, indicating that \task{} offers unique challenges that are not in E-mail summarization.






\section{Conclusion\label{sec:conclusion}}
We propose the \task\ task to summarize QA pairs in Community-based Question Answering. 
We develope a multi-stage annotation framework and created a benchmark \corpus{} for the \task\ task. Our multi-stage annotation framework decomposes a complex annotation task into three much simpler ones, thus allows higher annotation quality. We further compare a collection of extractive and abstractive summarization methods and establish a strong baseline method DedupLED for the \task\ task. Our experiment also confirms two key challenges, sentence-type transfer and duplication removal, towards the \task\ task. 

%




\section*{Limitations}

%
%
%
%

As we propose and tackle a challenging summarization task, the paper has certain limitations. First, our benchmark is in a single domain (E-commerce) in a single language (English), 
which not necessarily ensuring the generalizability for other domains and languages. 
Second, the quality of our annotations relies on the initial selection of seed QA pairs. As we discussed in the paper, we filtered high-quality seed QA pairs to minimize the risk. Nevertheless, it may not accurately replicate the summarization procedure by experts.
Third, we use rules and heuristics to ensure the quality of the free-text annotation. Despite being able to detect and eliminate a significant ratio of low-quality annotation, our rules and heuristics do not provide perfect guarantee, meaning that \corpus{} may still contain noisy and low-quality annotations.  
With those limitations, we still believe that the paper and the benchmark are benefitial for the community to take a step beyond the scope of existing summarization tasks. 

\section*{Ethics Statement}
For the annotation tasks, we paid \$10 hourly wage for the crowd workers on MTurk and Appen (Steps 1 and 3) and $\$15$ to $\$30$ hourly wage for the Upwork contractors (Step 2), making sure to pay higher than the minimum wage in the U.S. (i.e., \$7.25 per hour). 
Our \corpus{} is based on the publicly available Amazon QA dataset. To our knowledge, the dataset does not contain any harmful content.
\bibliography{anthology,custom}

\begin{thebibliography}{29}
\expandafter\ifx\csname natexlab\endcsname\relax\def\natexlab#1{#1}\fi

\bibitem[{Amplayo et~al.(2022)Amplayo, Bražinskas, Suhara, Wang, and
  Liu}]{OpinionSummarizationSIGIR2022Tutorial}
Reinald~Kim Amplayo, Arthur Bražinskas, Yoshi Suhara, Xiaolan Wang, and Bing
  Liu. 2022.
\newblock \href {https://doi.org/10.48550/ARXIV.2206.01543} {Beyond opinion
  mining: Summarizing opinions of customer reviews}.

\bibitem[{Beltagy et~al.(2020)Beltagy, Peters, and
  Cohan}]{beltagy2020longformer}
Iz~Beltagy, Matthew~E Peters, and Arman Cohan. 2020.
\newblock Longformer: The long-document transformer.
\newblock \emph{arXiv preprint arXiv:2004.05150}.

\bibitem[{Chan et~al.(2012)Chan, Zhou, Wang, and
  Chua}]{chan-etal-2012-community}
Wen Chan, Xiangdong Zhou, Wei Wang, and Tat-Seng Chua. 2012.
\newblock \href {https://aclanthology.org/P12-1061} {Community answer
  summarization for multi-sentence question with group {L}1 regularization}.
\newblock In \emph{Proceedings of the 50th Annual Meeting of the Association
  for Computational Linguistics (Volume 1: Long Papers)}, pages 582--591, Jeju
  Island, Korea. Association for Computational Linguistics.

\bibitem[{Chen and Bansal(2018)}]{chen2018fast}
Yen-Chun Chen and Mohit Bansal. 2018.
\newblock Fast abstractive summarization with reinforce-selected sentence
  rewriting.
\newblock \emph{arXiv preprint arXiv:1805.11080}.

\bibitem[{Chowdhury and Chakraborty(2019)}]{CQASumm}
Tanya Chowdhury and Tanmoy Chakraborty. 2019.
\newblock \href {https://doi.org/10.1145/3297001.3297004} {{CQASUMM}: Building
  references for community question answering summarization corpora}.
\newblock In \emph{Proceedings of the ACM India Joint International Conference
  on Data Science and Management of Data}, CoDS-COMAD '19, page 18–26, New
  York, NY, USA. Association for Computing Machinery.

\bibitem[{Chowdhury et~al.(2021)Chowdhury, Kumar, and
  Chakraborty}]{chowdhury2021neural}
Tanya Chowdhury, Sachin Kumar, and Tanmoy Chakraborty. 2021.
\newblock Neural abstractive summarization with structural attention.
\newblock In \emph{Proceedings of the Twenty-Ninth International Conference on
  International Joint Conferences on Artificial Intelligence}, pages
  3716--3722.

\bibitem[{Deng et~al.(2020{\natexlab{a}})Deng, Lam, Xie, Chen, Li, Yang, and
  Shen}]{Deng:2020:AAAI}
Yang Deng, Wai Lam, Yuexiang Xie, Daoyuan Chen, Yaliang Li, Min Yang, and Ying
  Shen. 2020{\natexlab{a}}.
\newblock \href {https://ojs.aaai.org/index.php/AAAI/article/view/6266} {Joint
  learning of answer selection and answer summary generation in community
  question answering}.
\newblock In \emph{The Thirty-Fourth {AAAI} Conference on Artificial
  Intelligence, {AAAI} 2020}, pages 7651--7658. {AAAI} Press.

\bibitem[{Deng et~al.(2020{\natexlab{b}})Deng, Zhang, Li, Yang, Lam, and
  Shen}]{Deng:2020:SIGIR}
Yang Deng, Wenxuan Zhang, Yaliang Li, Min Yang, Wai Lam, and Ying Shen.
  2020{\natexlab{b}}.
\newblock \href {https://doi.org/10.1145/3397271.3401208} {\emph{Bridging
  Hierarchical and Sequential Context Modeling for Question-Driven Extractive
  Answer Summarization}}, page 1693–1696. Association for Computing
  Machinery, New York, NY, USA.

\bibitem[{Devlin et~al.(2019)Devlin, Chang, Lee, and
  Toutanova}]{devlin2019bert}
Jacob Devlin, Ming-Wei Chang, Kenton Lee, and Kristina Toutanova. 2019.
\newblock Bert: Pre-training of deep bidirectional transformers for language
  understanding.
\newblock In \emph{Proceedings of the 2019 Conference of the North American
  Chapter of the Association for Computational Linguistics: Human Language
  Technologies, Volume 1 (Long and Short Papers)}, pages 4171--4186.

\bibitem[{Erkan and Radev(2004)}]{erkan2004lexrank}
G{\"u}nes Erkan and Dragomir~R Radev. 2004.
\newblock Lexrank: Graph-based lexical centrality as salience in text
  summarization.
\newblock \emph{Journal of artificial intelligence research}, 22:457--479.

\bibitem[{Fabbri et~al.(2021{\natexlab{a}})Fabbri, Rahman, Rizvi, Wang, Li,
  Mehdad, and Radev}]{fabbri-etal-2021-convosumm}
Alexander Fabbri, Faiaz Rahman, Imad Rizvi, Borui Wang, Haoran Li, Yashar
  Mehdad, and Dragomir Radev. 2021{\natexlab{a}}.
\newblock \href {https://doi.org/10.18653/v1/2021.acl-long.535} {{C}onvo{S}umm:
  Conversation summarization benchmark and improved abstractive summarization
  with argument mining}.
\newblock In \emph{Proceedings of the 59th Annual Meeting of the Association
  for Computational Linguistics and the 11th International Joint Conference on
  Natural Language Processing (Volume 1: Long Papers)}, pages 6866--6880,
  Online. Association for Computational Linguistics.

\bibitem[{Fabbri et~al.(2021{\natexlab{b}})Fabbri, Wu, Iyer, Li, and
  Diab}]{AnswerSumm}
Alexander~R. Fabbri, Xiaojian Wu, Srini Iyer, Haoran Li, and Mona Diab.
  2021{\natexlab{b}}.
\newblock \href {https://doi.org/10.48550/ARXIV.2111.06474} {Answersumm: A
  manually-curated dataset and pipeline for answer summarization}.

\bibitem[{Feigenblat et~al.(2021)Feigenblat, Gunasekara, Sznajder, Joshi,
  Konopnicki, and Aharonov}]{feigenblat-etal-2021-tweetsumm-dialog}
Guy Feigenblat, Chulaka Gunasekara, Benjamin Sznajder, Sachindra Joshi, David
  Konopnicki, and Ranit Aharonov. 2021.
\newblock \href {https://doi.org/10.18653/v1/2021.findings-emnlp.24}
  {{TWEETSUMM} - a dialog summarization dataset for customer service}.
\newblock In \emph{Findings of the Association for Computational Linguistics:
  EMNLP 2021}, pages 245--260, Punta Cana, Dominican Republic. Association for
  Computational Linguistics.

\bibitem[{Iso et~al.(2022)Iso, Wang, Angelidis, and
  Suhara}]{iso-etal-2022-comparative}
Hayate Iso, Xiaolan Wang, Stefanos Angelidis, and Yoshihiko Suhara. 2022.
\newblock \href {https://doi.org/10.18653/v1/2022.findings-acl.261}
  {Comparative opinion summarization via collaborative decoding}.
\newblock In \emph{Findings of the Association for Computational Linguistics:
  ACL 2022}, pages 3307--3324, Dublin, Ireland. Association for Computational
  Linguistics.

\bibitem[{Khalman et~al.(2021)Khalman, Zhao, and
  Saleh}]{khalman-etal-2021-forumsum-multi}
Misha Khalman, Yao Zhao, and Mohammad Saleh. 2021.
\newblock \href {https://doi.org/10.18653/v1/2021.findings-emnlp.391}
  {{F}orum{S}um: A multi-speaker conversation summarization dataset}.
\newblock In \emph{Findings of the Association for Computational Linguistics:
  EMNLP 2021}, pages 4592--4599, Punta Cana, Dominican Republic. Association
  for Computational Linguistics.

\bibitem[{Lebanoff et~al.(2019)Lebanoff, Song, Dernoncourt, Kim, Kim, Chang,
  and Liu}]{lebanoff2019scoring}
Logan Lebanoff, Kaiqiang Song, Franck Dernoncourt, Doo~Soon Kim, Seokhwan Kim,
  Walter Chang, and Fei Liu. 2019.
\newblock Scoring sentence singletons and pairs for abstractive summarization.
\newblock \emph{arXiv preprint arXiv:1906.00077}.

\bibitem[{Lewis et~al.(2019)Lewis, Liu, Goyal, Ghazvininejad, Mohamed, Levy,
  Stoyanov, and Zettlemoyer}]{lewis2019bart}
Mike Lewis, Yinhan Liu, Naman Goyal, Marjan Ghazvininejad, Abdelrahman Mohamed,
  Omer Levy, Ves Stoyanov, and Luke Zettlemoyer. 2019.
\newblock Bart: Denoising sequence-to-sequence pre-training for natural
  language generation, translation, and comprehension.
\newblock \emph{arXiv preprint arXiv:1910.13461}.

\bibitem[{Lin(2004)}]{lin2004rouge}
Chin-Yew Lin. 2004.
\newblock Rouge: A package for automatic evaluation of summaries.
\newblock In \emph{Text summarization branches out}, pages 74--81.

\bibitem[{Liu and Lapata(2019)}]{liu2019text}
Yang Liu and Mirella Lapata. 2019.
\newblock Text summarization with pretrained encoders.
\newblock In \emph{Proceedings of the 2019 Conference on Empirical Methods in
  Natural Language Processing and the 9th International Joint Conference on
  Natural Language Processing (EMNLP-IJCNLP)}, pages 3730--3740.

\bibitem[{Liu et~al.(2008)Liu, Li, Cao, Lin, Han, and
  Yu}]{liu2008understanding}
Yuanjie Liu, Shasha Li, Yunbo Cao, Chin-Yew Lin, Dingyi Han, and Yong Yu. 2008.
\newblock Understanding and summarizing answers in community-based question
  answering services.
\newblock In \emph{Proceedings of the 22nd International Conference on
  Computational Linguistics (COLING 2008)}, pages 497--504.

\bibitem[{Louviere et~al.(2015)Louviere, Flynn, and Marley}]{louviere2015best}
Jordan~J Louviere, Terry~N Flynn, and Anthony Alfred~John Marley. 2015.
\newblock \emph{Best-worst scaling: Theory, methods and applications}.
\newblock Cambridge University Press.

\bibitem[{McAuley and Yang(2016)}]{mcauley2016addressing}
Julian McAuley and Alex Yang. 2016.
\newblock Addressing complex and subjective product-related queries with
  customer reviews.
\newblock In \emph{Proceedings of the 25th International Conference on World
  Wide Web}, pages 625--635.

\bibitem[{Oved and Levy(2021)}]{oved-levy-2021-pass}
Nadav Oved and Ran Levy. 2021.
\newblock \href {https://doi.org/10.18653/v1/2021.acl-long.30} {{PASS}:
  Perturb-and-select summarizer for product reviews}.
\newblock In \emph{Proceedings of the 59th Annual Meeting of the Association
  for Computational Linguistics and the 11th International Joint Conference on
  Natural Language Processing (Volume 1: Long Papers)}, pages 351--365, Online.
  Association for Computational Linguistics.

\bibitem[{Raffel et~al.(2020)Raffel, Shazeer, Roberts, Lee, Narang, Matena,
  Zhou, Li, Liu et~al.}]{raffel2020exploring}
Colin Raffel, Noam Shazeer, Adam Roberts, Katherine Lee, Sharan Narang, Michael
  Matena, Yanqi Zhou, Wei Li, Peter~J Liu, et~al. 2020.
\newblock Exploring the limits of transfer learning with a unified text-to-text
  transformer.
\newblock \emph{J. Mach. Learn. Res.}, 21(140):1--67.

\bibitem[{Wan and McAuley(2016)}]{wan2016modeling}
Mengting Wan and Julian McAuley. 2016.
\newblock Modeling ambiguity, subjectivity, and diverging viewpoints in opinion
  question answering systems.
\newblock In \emph{2016 IEEE 16th international conference on data mining
  (ICDM)}, pages 489--498. IEEE.

\bibitem[{Zhang et~al.(2020)Zhang, Zhao, Saleh, and Liu}]{zhang2020pegasus}
Jingqing Zhang, Yao Zhao, Mohammad Saleh, and Peter Liu. 2020.
\newblock Pegasus: Pre-training with extracted gap-sentences for abstractive
  summarization.
\newblock In \emph{International Conference on Machine Learning}, pages
  11328--11339. PMLR.

\bibitem[{Zhang et~al.(2021)Zhang, Celikyilmaz, Gao, and
  Bansal}]{zhang2021emailsum}
Shiyue Zhang, Asli Celikyilmaz, Jianfeng Gao, and Mohit Bansal. 2021.
\newblock Emailsum: Abstractive email thread summarization.
\newblock \emph{arXiv preprint arXiv:2107.14691}.

\bibitem[{Zhang et~al.(2019)Zhang, Kishore, Wu, Weinberger, and
  Artzi}]{zhang2019bertscore}
Tianyi Zhang, Varsha Kishore, Felix Wu, Kilian~Q Weinberger, and Yoav Artzi.
  2019.
\newblock Bertscore: Evaluating text generation with bert.
\newblock \emph{arXiv preprint arXiv:1904.09675}.

\bibitem[{Zhu et~al.(2020)Zhu, Xu, Zeng, and Huang}]{zhu2020hierarchical}
Chenguang Zhu, Ruochen Xu, Michael Zeng, and Xuedong Huang. 2020.
\newblock A hierarchical network for abstractive meeting summarization with
  cross-domain pretraining.
\newblock \emph{arXiv preprint arXiv:2004.02016}.

\end{thebibliography}
\bibliographystyle{acl_natbib}



\end{document}